# A Vision-based System for Traffic Anomaly Detection using Deep Learning and Decision Trees

Armstrong Aboah[1*], Maged Shoman[1*], Vishal Mandal[1], Sayedomidreza Davami[2], Yaw Adu-Gyamfi[1#], Anuj Sharma[2]

[1]Department of Civil and Environmental Engineering, University of Missouri-Columbia
[2]Department of Civil and Environmental Engineering, Iowa State University

*Abstract*— **Any intelligent traffic monitoring system must be able to detect anomalies such as traffic accidents in real time. In this paper, we propose a Decision-Tree - enabled approach powered by Deep Learning for extracting anomalies from traffic cameras while accurately estimating the start and end time of the anomalous event. Our approach included creating a detection model, followed by anomaly detection and analysis. YOLOv5 served as the foundation for our detection model. The anomaly detection and analysis step entail traffic scene background estimation, road mask extraction, and adaptive thresholding. Candidate anomalies were passed through a decision tree to detect and analyze final anomalies. The proposed approach yielded an F1 score of 0.8571, and an S4 score of 0.5686, per the experimental validation.**

## 1. Introduction

Advancement in consumer-level technologies such as video cameras has greatly improved traffic monitoring systems. In recent years, various State Traffic Management Centers (TMCs) rely on live CCTV footage to coordinate and provide appropriate responses to various highway traffic incidents. The current generation of traffic monitoring systems are however costly to maintain because they are manually operated. The lack of automation also leads to low incident detection rates and response times. The size, resolution and speed at which data from traffic monitoring systems arrive can also be overwhelming for traffic operators. Thus, there is a need to develop scalable applications which can quickly ingest traffic condition data from cameras and extract information relevant for coordinating and responding to traffic incidents or anomalies. This paper presents a methodology that effectively and efficiently identifies anomalies or incidents in CCTV footage using state-of-the-art deep learning and computer vision-based models.

Since traffic monitoring systems must operate in real-time and under varying traffic and weather conditions, automated, vision-based traffic anomaly detection is a difficult problem to solve. Traditional vision-based systems are often hampered by typical traffic scene features such as heavy occlusion and poor video quality [1, 2]. Numerous studies have been geared towards improving the accuracy of anomaly detection systems. For instance, the use of graphical models to model vehicular motion parameters, which are critical for detecting anomaly scenes, has been investigated by [2]. Elahi et al. used a Parzen probabilistic neural network trained on video data to track traffic anomalies on the road [3]. Thajchayapong et al. [4] proposed a multi-resolution anomaly classification algorithm based on the productivities of neural networks to predict anomalous traffic conditions. Similarly, in [5], a probabilistic neural network is used to classify various traffic conditions into multiple categories. Several studies on traffic anomalies have proposed incident precursor algorithms that track hazardous traffic conditions [5-7]. While the preceding research was successful in ideal conditions, it relied on detectors and probe datasets rather than video data.

In this study, the authors developed a framework for detecting traffic anomalies in video data. The proposed methodology relies on an augmented annotation pipeline which pre-annotate the training dataset using an object-detection model trained on the COCO dataset. Annotations are subsequently used to build a vehicle detection model using the YOLOv5 network. Next, we estimate the background of each traffic video by computing the median of frames randomly sampled from a uniform distribution over a thirty second period. Vehicle detections on extracted backgrounds are classified as anomaly candidates. Factors such as vehicle detection size, likelihood, and road feature masks were used to construct a decision tree to eliminate false anomalies. The start and end of an anomaly were computed by superimposing detections from anomaly candidates and their foreground detections.

The rest of the paper is organized as follows. Section two provides a review of relevant literature. The data used for this study is present in Section three. Section four presents the methodology employed by this study. Section five presents a discussion of results from the model development. Finally, Section six presents a summary of the research, the conclusions drawn from the results, and recommendations for future research.

## 2. Related Work

Detecting anomaly with typical aberration in vehicle scene entities remains an important subdomain of traffic behavior modeling [8]. Due to the accessibility to traffic video scenes, there has been an upsurge of research in the areas of video analysis and anomaly detection [9-13]. Since most computer vision models typically analyze general traffic scenes and separate the abnormal from normal traffic events, methods such as Markov model [16-17], Markov Random Field [11-12] and Sparse Reconstruction [20-22] have enjoyed some successes. However, with the advent of deep learning, there has been significant improvements in detecting traffic anomalies. Therefore, a clear majority of studies deploy deep neural networks to detect them. Li et al. in [15] proposed a multi-granularity vehicle tracking technique with modularized elements, where it uses Faster R-CNN, a deep learning framework to build its object detection module. Likewise, its modularized element consists of the object detector, background modeler, mask extractor and tracker. Their method used both box and pixel-level tracking strategy to ameliorate anomaly prediction results. The pixel level tracking in [15] was inspired by the winning solution of the 2019 AI City Challenge [26]. Needless to say, the combination of both those strategies followed by the backtracking optimization technique helped [15] attain first rank in the anomaly detection track of the 2020 NVIDIA AI City Challenge [29].

Generally, most anomaly detection methods are supervised with the exception of some that focus on unsupervised techniques. Zhao et al. in [14] proposed an unsupervised anomaly detection framework through information gained from vehicle trajectories. Their method obtained superior results deploying a multi-object tracker to mitigate the effects of false detections caused by the detector. Mandal et al. in [23] used a pre-trained YOLO network and feature tracker to detect traffic anomalies such as stopped vehicles and roadside accidents. An anomaly detection system in [24] leverages a YOLO based object detector, coupled with post processing modules to predict stationary vehicles through nearest neighbors and K-means clustering technique. Although their nearest neighbor and clustering technique levied extensive training requirements, training on anomalous traffic video feeds could have extracted superior performances. Bai et al. in [26] proposed an anomaly detection system consisting of the spatial temporal matrix discriminating module along with the background modeler and perspective detection module. The spatial temporal matrix module used in their study transformed the analysis of strip trajectory into the study of spatial position which furnished accurate start and stop times, and an improved anomaly detection score leading to a first place finish on the 2019 NVIDIA AI City Challenge leaderboard [31].

In the current study, the authors employed state-of-the-art YOLO object detection framework and focused on a more heuristic approach around post-processing modules to detect anomalies. Unlike some studies [14, 25] that deploy vehicle tracking algorithms, our proposed approach circumvents the use of a tracker especially since the clear majority of vehicles in a traffic scene would have made tracking individual vehicles difficult and computationally infeasible. It is worth mentioning that most of the winning teams from the 2018-2020 NVIDIA AI City Challenge [29-31] emphasized on background image segmentation and improving vehicle detection along with some post processing modules. Inspired by these earlier solutions, our approach uses a simple, yet efficient framework for background estimation and road segmentation. A decision tree approach is also adopted for characterizing anomalies using information from detections on foreground and background images.

## 3. Data

The data used to train and test the proposed anomaly detection algorithm was provided by NVIDIA AI CITY CHALLENGE 2021. For Track 4 (Traffic Anomaly Detection), the data consists of: 100 videos for training and 150 videos for testing with an average length of 15 minutes, 30 fps and a resolution of 410p. Each video presents a unique challenge since they are a mix of road types, diverse camera angles, lighting and weather conditions. The main objective is to detect an anomaly, defined as vehicle stoppage due to a crash or stall.

## 4. Proposed Methodology

The proposed methodology can be broken down into various steps, as shown in Figure 1. The videos are first sorted using an automated video sorting system. Following that, a concurrent process of detecting foreground objects and estimating background features is carried out. Next, background images are passed through the vehicle object detector to flag potential anomalies. Finally, a decision tree is used to detect and isolate false anomalies based on predefined rules. The anomaly start and end times are calculated by superimposing the foreground and anomaly detections. A detailed description of each step is provided in the following sections.

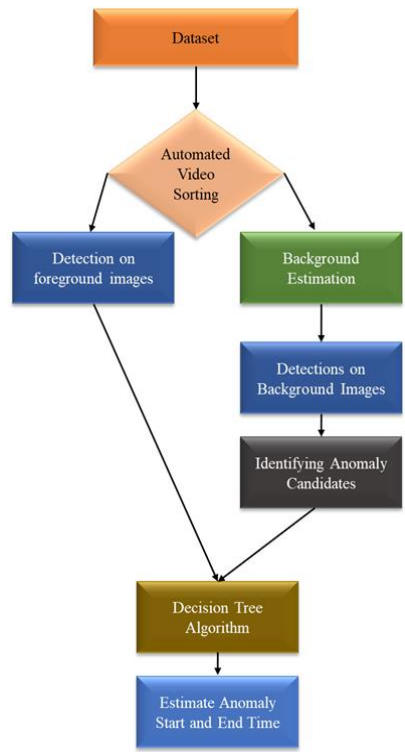

Figure 1: Flow chart of the full methodology

### 4.1. Vehicle Detection Model

This section involves using YOLOv5 to build a detection model and developing an augmented annotation system for the labeling of the training dataset.

### 4.1.1. YOLOv5

YOLOv5 [27] is the latest iteration to YOLO series and a state-of-the-art single stage object detection algorithm. The

YOLOv5 network consists of three main pieces viz. Backbone, Neck and Head. The Backbone consists of a convolutional neural network that bundles and forms image representational features at contrasting granularities. The architecture's neck consists of a series of layers which blends and integrates image representational features to proceed further with prediction. Similarly, the head utilizes features from the neck and gets hold of box and class prediction functionality. CSPDarknet53 backbone within YOLOv5 contains 29 convolutional layers $3 \times 3$, receptive field size of $725 \times 725$ and altogether 27.6 M parameters. Besides, the SPP block attached over YOLO's CSPDarknet53 expands the proportion of receptive fields without influencing its operating speed. Likewise, the feature aggregation is performed through PANet by exploiting different levels of backbone. YOLOv5 pushes state-of-the-art by using features such as the weighted-residual-connections, cross-stage-partial-connections, cross mini-batch, normalization and self-adversarial training, making it exceptionally efficient. In the current study, we trained and deployed our YOLOv5 model on the PyTorch [28] framework. To further accomplish the task of vehicle detection, the YOLOv5 model is fine-tuned by adjusting to the following hyperparameters: batch-size 64, the optimizer weight decay value of 0.0005, setting the initial learning rate of 0.01 and keeping the momentum at 0.937.

YOLOv5 [27] is the latest iteration to YOLO series and a state-of-the-art single stage object detection algorithm. The YOLOv5 network consists of three main pieces viz. Backbone, Neck and Head. The Backbone consists of a convolutional neural network that bundles and forms image representational features at contrasting granularities. The architecture's neck consists of a series of layers which blends and integrates image representational features to proceed further with prediction. Similarly, the head utilizes features from the neck and gets hold of box and class prediction functionality. CSPDarknet53 backbone within YOLOv5 contains 29 convolutional layers $3 \times 3$, receptive field size of $725 \times 725$ and altogether 27.6 M parameters. Besides, the SPP block attached over YOLO's CSPDarknet53 expands the proportion of receptive fields without influencing its operating speed. Likewise, the feature aggregation is performed through PANet by exploiting different levels of backbone. YOLOv5 pushes state-of-the-art by using features such as the weighted-residual-connections, cross-stage-partial-connections, cross mini-batch, normalization and self-adversarial training, making it exceptionally efficient. In the current study, we trained and deployed our YOLOv5 model on the PyTorch [28] framework. To further accomplish the task of vehicle detection, the YOLOv5 model is fine-tuned by adjusting to the following hyperparameters: batch-size 64, the optimizer weight decay value of 0.0005, setting the initial learning rate of 0.01 and keeping the momentum at 0.937.

#### 4.1.2. Augmented Annotation

The main idea here is to reduce annotation time by first automatically generating annotations from an already existing model, followed by manual verification and correction. In this work, the training images were pre-annotated using a YOLOv5 network trained on the ImageNet dataset. with an existing model Objects that were wrongly annotated by the network were manually corrected. The corrected labels were used to re-train a new YOLOv5 model. The process is repeated until the model accuracy converges on the test dataset. Figure 2 summarizes the augmented annotation framework.

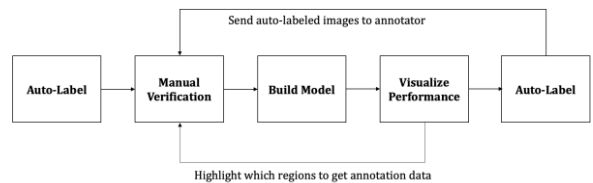

Figure 2: Augmented Annotation

#### 4.2.1. Video Sorting

Videos are first automatically sorted based on the road type (freeway or intersection), weather condition (snow), and time of day (night vs day). A road is classified as an intersection or interchange if more than two directions are detected whereas a road is designated as a freeway or two-lane road when two directions are detected. The number of directions is estimated from the number of unique vehicle trajectories per scene. Figure 4a for examples shows a detected intersection as the number of unique trajectories are more than 2. The frequency distribution of image pixels was used to sort videos by time of day and weather condition. Night videos typically have a unimodal distribution with a peak closest to 0–50-pixel value.

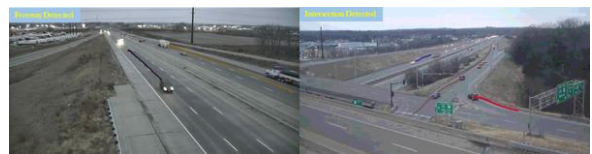

Figure 3a. Road type classification

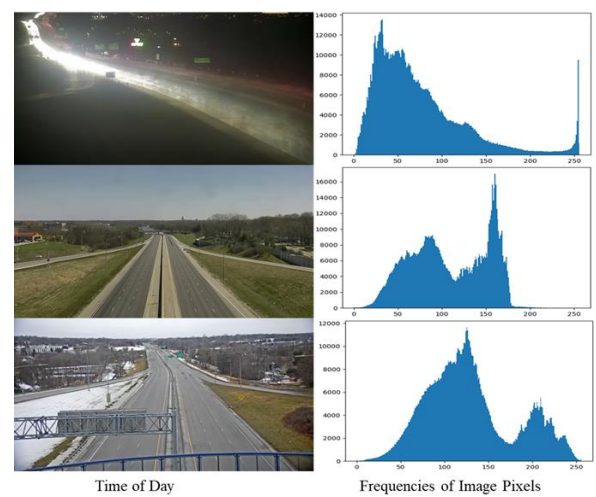

Time of Day        Frequencies of Image Pixels

Figure 3b. Sorting videos by time of the day and weather condition using adaptive thresholding

A bimodal distribution centered around pixel values ranging between 100 - 150 is observed for daylight videos.

Videos captured under snow conditions also have a bimodal distribution but are centered around 200 - 250-pixel values. An average histogram of all frames in a video was computed and thresholded (using the aforementioned thresholds) to sort videos into day, night and snowy as shown in Figure 3b.

### 4.2. Anomaly Detection and Analysis

The anomaly detection process has three main components: a background estimator, road mask extractor and a decision tree. The sections below explain each step in detail.

#### 4.2.1. Background Estimation

The background of an ideal video is estimated by first, randomly sampling frames within a 30 - second period, followed by calculating the median of 10% of all frames in the sample. By random sampling and taking a median of a subset of images, we are able to eliminate the effect of short-term video resolution changes such as zooms, pixelation, etc. The frame sampling periods were varied based on the outputs from the video sorting algorithm. Backgrounds for intersections, night-time and videos capturing snow conditions were estimated at 5-minute intervals as compared to 30 seconds for an ideal video. Figures 4 a-c shows example background features extracted at different time intervals.

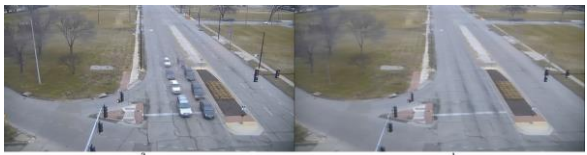

Figure 4a. left - Background image using median frames over 30 seconds, right- Background image using median frames over 180 seconds

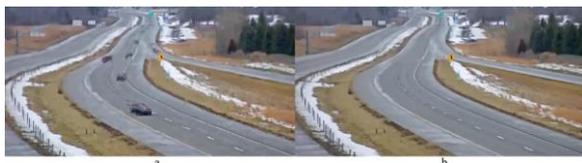

Figure 4b. left - Background image for a pixelated video using median frames over 30 seconds, right - Background image using median frames over 180 seconds

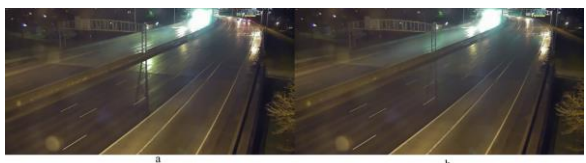

Figure 4c. left - Background image using median frames over 30 seconds, right - Background image using median frames over 180 seconds

The main reasons for varying the time interval for generating background images includes the following. 1) Vehicles stopped at a stop sign or at a traffic light may appear to be stationary objects if background images are generated for a short period of time (30 seconds), this problem can be avoided; however, if a long period of time is used, as shown in Figure 4a. 2) Highly pixelate and nigh videos may also generate false stationary objects in the background if the background images are generated over a shorter period of time, as shown in Figure 4b-c.

#### 4.2.2. Identifying Anomaly Candidates

Candidate anomalies are extracted by passing each background image through the YOLOv5 object detection network developed. Any vehicle detected in the background is considered as an anomaly candidate. Figure 5 below shows examples of vehicular objects detected on background images.

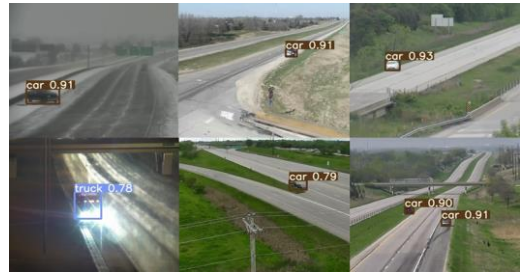

Figure 5. Background detections

On background images where parking lots are present, the preceding step is likely to flag parked vehicles as anomalies. To filter out these false anomaly detections, we used an adaptive image thresholding technique (designated by the equation below) to generate road masks from the background images as shown in Figure 6. The equation for background image thresholding is shown below. K1 and K2 are selected based on the outputs from our video sorting algorithm.

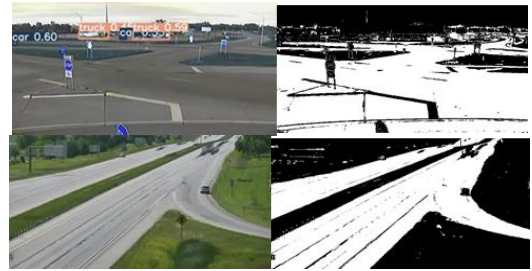

Figure 6. Road mask generation for filtering out false anomaly candidates.

$$(\mu - K1\ \sigma)\ /\ K\_2 \leq T \leq (\mu + K1\ \sigma)\ /\ (K1 + K2)$$

Candidate anomalies whose bounding boxes do not intersect with road masks (white regions) are flagged as false anomalies. The remaining candidates are passed through a decision tree to confirm and finalize anomaly detections.

The full decision tree algorithm is presented in Figure 7. It takes in both video foreground and background detections as inputs. If the background detection score and its area are greater than a pre-defined threshold, we compute an IOU between the detected anomaly candidate and the foreground detections. The frequency of overlapping foreground and background detections are then used to decide if an anomaly is present or not. The first and last instance when the background and foreground detections overlap is used to estimate the start and end time of the anomaly.

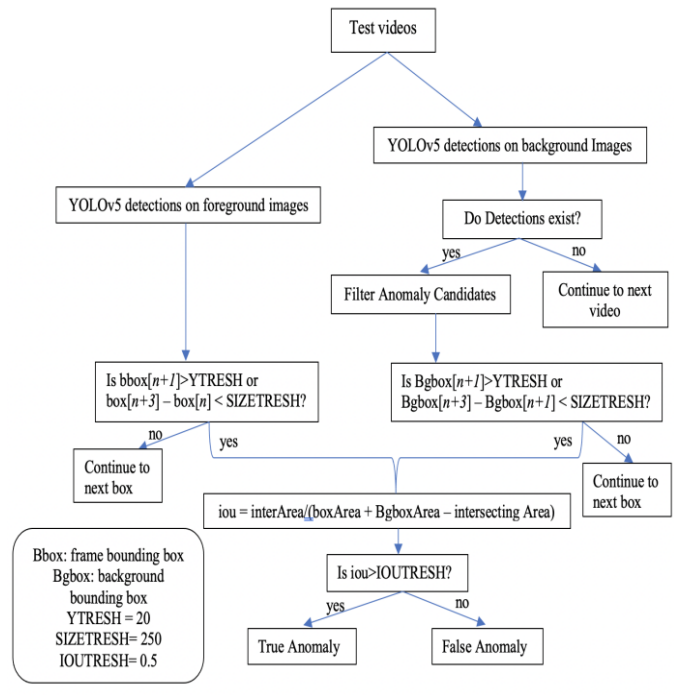

Figure 7: Anomaly detection decision tree

## 5. Results and Discussion

Our model is evaluated based on its capability to identify videos with anomalies measured by the F1score and the start and end time of the identified anomaly measured by the root mean square error (RMSE). Leaderboard ranking for submission is based on the S4 score, calculated using the following formula:

$$S4 = F1 \times (1 - NRMSE)$$

NRMSE is the normalized root mean square error that normalizes RMSE to scores between 0 and 1. A perfect score of 1 is achieved if the anomaly is detected within 10 seconds from the ground truth, while a value of 0 is achieved if the RMSE is at a maximum of 300 for detections at 5mins or higher from the ground truth.

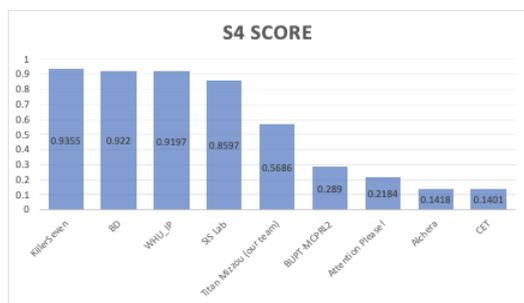

Figure 8: Leaderboard Rank

As shown in Figure 8, our traffic anomaly detection model was ranked 5th place on the leaderboard with an S4 score of 0.5686. Our F1 score of 0.8571 indicates that our model was successful in identifying anomalies across most videos and an RMSE of 101.0071 reflects our absolute model fit to the data where due to the precision limitations of the dataset used for training, the detector was not able to perfectly determine objects that are too small or too far away from the camera when they first appeared as an anomaly.

## 6. Conclusion and Recommendation

The rapid advancement in the field of machine learning and high-performance computing has significantly amplified the scope of automated anomaly prediction systems. In this research, we implemented a deep learning-based model for traffic scene anomaly detection. The F1, RMSE and S4 scores for the model were found to be 0.8571, 101.0071, and 0.5686, respectively. It became evident through the results that the model developed could capture anomalies located near the camera but had issues capturing distant anomalies. Factors such as video pixelation and traffic intersections also contributed to S4 scores. Different approaches, such as video sorting and anomaly candidate filtering, were used to improve the effectiveness of the proposed framework for anomaly detection. Future work would focus on considering the combination of the IOU tracker with more robust tracking algorithms, thereby increasing the annotation database with distant vehicular objects.